\documentclass{article}

    \PassOptionsToPackage{numbers, compress}{natbib}

    \usepackage[final]{neurips_2020}

\usepackage[utf8]{inputenc} 
\usepackage[T1]{fontenc}    
\usepackage{hyperref}       
\usepackage{url}            
\usepackage{booktabs}       
\usepackage{amsfonts}       
\usepackage{nicefrac}       
\usepackage{microtype}      
\usepackage{amsmath}
\usepackage{subfig}
\usepackage{graphicx}
\usepackage{lipsum}
\usepackage{graphicx}

\usepackage{xcolor}
\usepackage{multirow}
\usepackage{wrapfig}
\usepackage{array}
\usepackage{soul}
\usepackage{caption}
\usepackage{bm}
\usepackage{hyperref}
\hypersetup{
    colorlinks=true,
    linkcolor=red,
    filecolor=magenta,      
    urlcolor=cyan,
}

\title{Cross-directional Feature Fusion Network for Building Damage Assessment from Satellite Imagery}

\author{%
  Yu Shen$^{1,2}$,
   Sijie Zhu$^1$, Taojiannan Yang$^1$, Chen Chen$^1$ \\
 ${}^1$University of North Carolina at Charlotte, ${}^2$Nanjing University of Science and Technology\\
  \texttt{shenyu@njust.edu.cn}, \texttt{\{szhu3,tyang30,chen.chen\}@uncc.edu} \\
}

\begin{document}

\maketitle

\begin{abstract}
Fast and effective responses are required when a natural disaster (e.g., earthquake, hurricane, etc.) strikes. Building damage assessment from satellite imagery is critical before an effective response is conducted. High-resolution satellite images provide rich information with pre- and post-disaster scenes for analysis. However, most existing works simply use pre- and post-disaster images as input without considering their correlations. In this paper, we propose a novel cross-directional fusion strategy to better explore the correlations between pre- and post-disaster images. Moreover, the data augmentation method CutMix is exploited to tackle the challenge of hard classes. The proposed method achieves state-of-the-art performance on a large-scale building damage assessment dataset -- xBD.

\end{abstract}

\section{Introduction}
\label{secIntro}
Natural disasters, such as earthquakes, floods and tsunami, cause serious social and economic devastation. When a natural disaster strikes, accurate and immediate responses are required in Humanitarian Assistance and Disaster Response (HADR) for saving thousands of lives \cite{sidrane2019machine, doshi2019firenet}. Before these responses, rescue planning and preparations are conducted based on the damage analysis \cite{weber2020detecting}. With the rapid development of remote sensing technology, high resolution satellite images are now available for damage analysis. Traditionally, these images of disaster areas are analyzed by experts, which might be time-consuming if the areas are large. Therefore, automatic information extraction from satellite images, such as building segmentation and damage assessment, is imperative under time-critical situations. 

Building damage assessment plays a pivotal role in HADR, which aims at predicting the building damage level for each pixel based on building segmentation. With a pair of pre- and post-disaster images, the extent of the damage to buildings can be learned by machine learning algorithms. Recently, deep learning-based methods have shown their effectiveness in building damage assessment. Xu et al. \cite{xu2019building} investigated the capability of convolutional neural networks (CNN) for building damage detection by identifying damaged and undamaged buildings. To evaluate the damage levels more precisely, Weber et al. \cite{weber2020building} considered building damage assessment as a semantic segmentation task. 

With a pair of pre- and post-disaster images for building damage assessment, a key question would be how to effectively model the correlations between these images? Unfortunately, there are only a few works have explored this direction.  
Hao et al. \cite{hao2020attention} simply concatenated the features from pre- and post-disaster images and fed them into non-local attention modules. 
Gupta et al. \cite{gupta2020rescuenet} developed a framework that uses the difference of pre- and post-disaster features as input of a network.

\begin{wrapfigure}{r}{0.35\textwidth}
    \vspace{-0.2cm}
    \includegraphics[width=\linewidth]{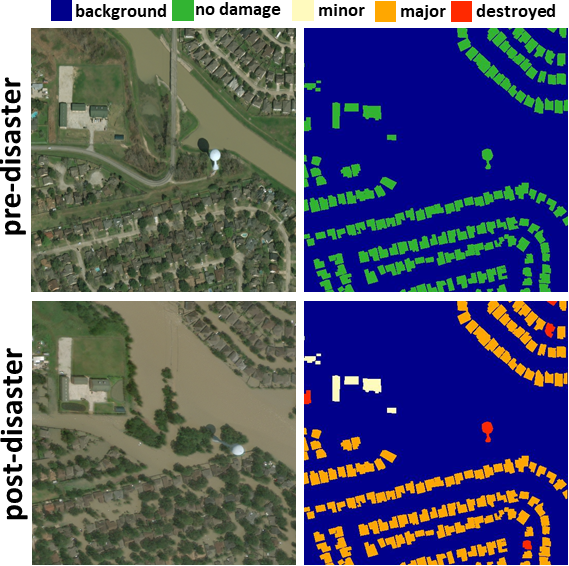} 
    \caption{A pair of images and their annotations from the xBD dataset.}
    \label{fig:figImage}
\end{wrapfigure}

Another challenge of building damage assessment from satellite imagery lies in the visual similarity between certain classes (e.g., \textit{no damage} and \textit{minor damage}). These classes are considered as hard classes. To better explain this problem, we use the xBD \cite{gupta2019xbd} dataset, which is the largest dataset for building damage assessment to date, as an example. Fig. \ref{fig:figImage} shows a pair of images from this dataset. 
Based on the visual observation, it is difficult to distinguish between classes such as \textit{no damage} and \textit{minor damage} due to high visual similarities. To further verify this observation, Table \ref{tabConfusion} reports the classification results of baseline method (ResNet-50) on xBD. From the classification confusion matrix, about 24.2\% of \emph{minor damage} are mis-classified as \emph{no damage}. 
\begin{wraptable}{r}{0.4\textwidth}
\vspace{-\intextsep}
  \caption{Classification confusion matrix (\%)  of ResNet-50 baseline on xBD test set.}
  \vspace{-\intextsep}
  \scriptsize
  \begin{center}
  \setlength\tabcolsep{3.0pt} 
  \renewcommand{\arraystretch}{0.6}
  \begin{tabular}{rccccc}
    \toprule
       Damage Level  & C0 & C1 & C2  & C3 & C4 \\
    \midrule
    Background (C0) &\textbf{98.6} &	0.9 &	0.2&	0.2 &	0.1 \\
    No damage (C1) & 7.1&	\textbf{88.7} &	3.2&	0.8&	0.1\\
    Minor Damage (C2) & 6.3 &	\textcolor{blue}{24.2}&	\textbf{60.0} &	9.2&	0.4\\
    Major Damage (C3)  & 3.0 &	6.6	& \textcolor{blue}{14.9}&	\textbf{73.1} &	2.4\\
    Destroyed (C4) & 5.5 &	2.4&	1.2&	8.9&	 \textbf{82.1} \\
    \bottomrule
  \end{tabular}
  \vspace{-\intextsep}
  \end{center}
 \label{tabConfusion}
\end{wraptable}
One effective strategy to cope with hard classes and improve the model performance is data augmentation \cite{shorten2019survey,cubuk2019autoaugment}. Data augmentation has been widely used as a pre-processing technique to artificially increase the size of dataset in segmentation tasks \cite{hernandez2018data, myronenko20183d, zhang2017mixup}. Recently, CutMix \cite{yun2019cutmix} is proposed as a new data augmentation technique, which generates a new image by combining two image samples, to enhance the generalization ability of neural networks. CutMix directly cuts and pastes image patches from one image to another, which can be easily used in segmentation tasks.

Motivated by the above observations, we introduce a two-stage U-Net \cite{ronneberger2015u} based framework that integrates pre- and post-disaster features for building damage assessment. First, a single U-Net is used for building segmentation. Then a two-branch U-Net is applied for damage assessment using the weights from building segmentation for fine tuning. In the network, a cross-directional fusion model is proposed to explore the correlations between features from pre- and post-disaster images. By leveraging channel-wise and spatial-wise correlations, the fused features can be further enhanced. Moreover, to tackle the hard classes problem, CutMix is employed for data augmentation. 
Specifically, we only apply CutMix to hard classes to make the network pay more attention to those classes, thereby learning more robust representations for hard classes. In the experiments, we show that this strategy yields superior classification performance over simply adopting CutMix for all classes. 

In summary, this work makes two key contributions. (1) We present a new framework that integrates pre- and post-disaster images for building damage assessment. The proposed cross-directional fusion model effectively aggregates the feature representations from two images. (2) We unveil the challenge of hard classes in building damage assessment and explore a data augmentation strategy CutMix to address this problem. The proposed framework achieves state-of-the-art performance on a large-scale building damage assessment benchmark -- xBD. 


\begin{figure}
  \centering
    \subfloat{
	\includegraphics[width=0.89\linewidth]{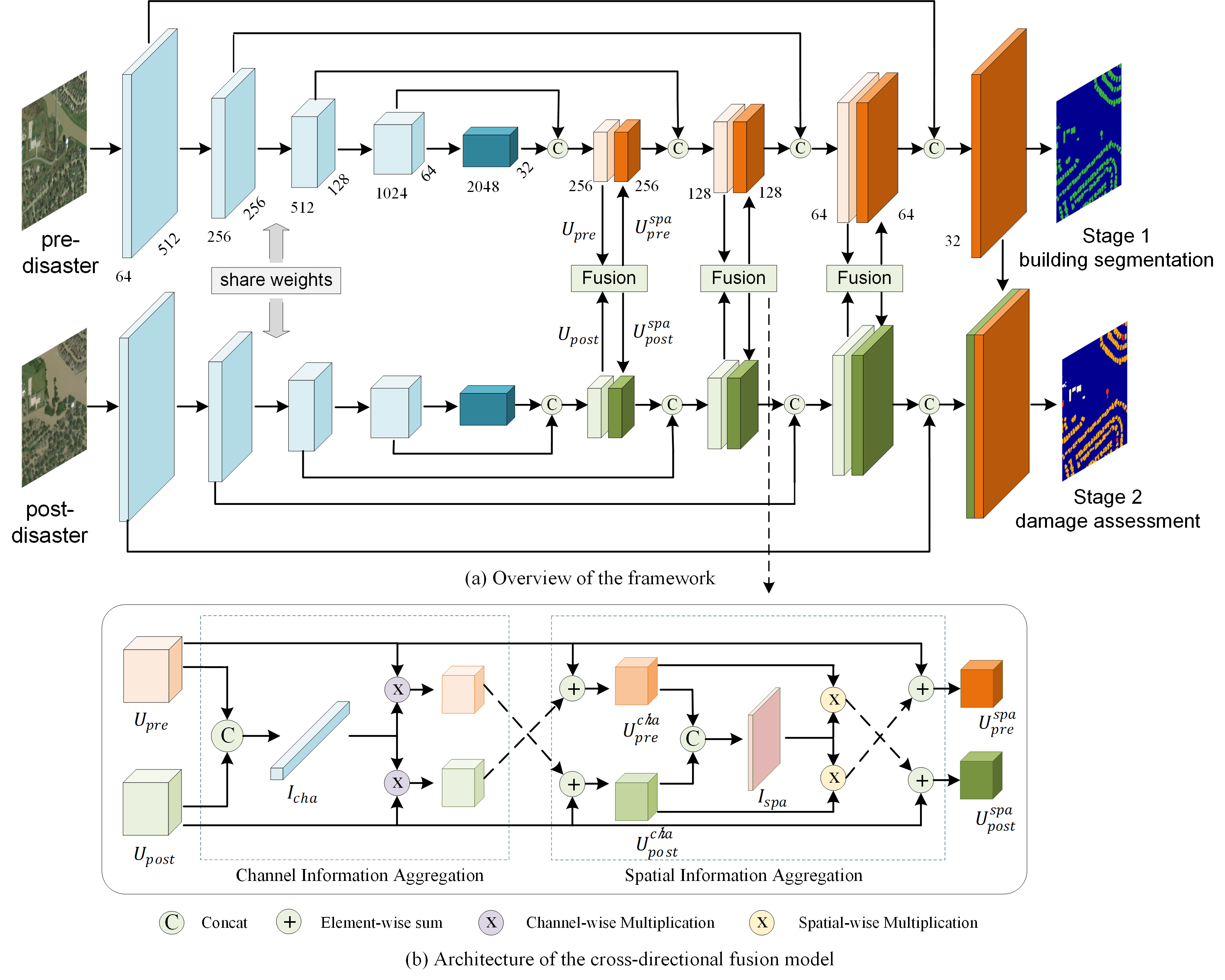} 
	} \\
  \caption{(a) Overall framework of the proposed method. In the building segmentation stage, only pre-disaster images and the upper U-Net branch are used. In the damage assessment stage, pre- and post disaster images are fed into the two network branches separately. (b) Architecture of the cross-directional fusion model.}
  \label{figFramework}
\end{figure}

\section{Proposed Method}
\label{secMethod}
\textbf{Overview.} 
As shown in Fig. \ref{figFramework}(a), the whole framework consists of two stages: building segmentation (stage 1) and damage assessment (stage 2). In stage 1, a single U-Net branch (i.e., the upper one) is used for building segmentation. This U-Net branch uses only pre-disaster images as input and produces segmentation masks of building objects. In stage 2, the pre- and post-disaster images are fed into the two network branches separately.
The weights are shared in the two-branch U-Net to reduce the computational cost. The network weights from stage 1 are used as initialization for network fine tuning in stage 2. To further enhance the feature representations, a cross-directional fusion model and CutMix data augmentation are utilized in the proposed framework.

\textbf{Cross-directional fusion model.} To further explore the correlations between pre- and post-disaster features, the proposed cross-directional fusion (CDF) model is added in the framework. Inspired by the squeeze and excitation (SE) block \cite{guha2018recalibrating}, the proposed CDF model focuses on recalibrating features from channel and spatial dimensions. 
Moreover, the channel and spatial information from  pre- and post-disaster features is aggregated together in a cross manner and then embedded in the network respectively. The model details are depicted in Fig. \ref{figFramework}(b). Let $U_{pre} \in\mathbb{R}^{C\times H\times W}$ and $U_{post}\in\mathbb{R}^{C\times H\times W}$ be the feature maps obtained from the two branches of U-Net respectively, the channel information can be extracted by
\begin{equation}
I_{cha} = \sigma(P([U_{pre}, U_{post}])), 
\label{eqAverageP}
\end{equation}
where $[U_{pre}, U_{post}]$ denotes the concatenation of feature maps, $P(\cdot)$ represents the global average pooling,  $\sigma(\cdot)$ is the sigmoid function and $I_{cha}$
is a feature vector of $C$ channels after dimension reduction from $2C$. Then the new features from two branches can be formulated as
\begin{equation}
\begin{aligned}
U_{pre}^{cha} = I_{cha}\ast U_{post} + U_{pre}  \quad \textrm{and} \quad
U_{post}^{cha} = I_{cha}\ast U_{pre} + U_{post},
\end{aligned}
\label{eqChannel}
\end{equation}
where $\ast$ denote the channel-wise multiplication between the input feature maps and vector $I_{cha}$. The output of channel feature fusion are then used in the spatial feature fusion. We concatenate $U_{pre}^{cha}$ and $U_{post}^{cha}$ and feed them into a $1\times1$ convolution $P_{conv}$ as follows
\begin{equation}
I_{spa} = \sigma(P_{conv}([U_{pre}^{cha}, U_{post}^{cha}])), 
\label{eqConv1}
\end{equation}
where $P_{conv}([U_{pre}^{cha}, U_{post}^{cha}]) \in \mathbb{R}^{1\times H \times W}$.
Then the output features are 
\begin{equation}
\begin{aligned}
U_{pre}^{spa} = I_{spa}\cdot U_{post}^{cha} + U_{pre}  \quad \textrm{and} \quad
U_{post}^{spa} = I_{spa}\cdot U_{pre}^{cha} + U_{post},
\end{aligned}
\end{equation}
where $I_{spa}\cdot U_{post}^{cha}$ and $I_{spa}\cdot U_{pre}^{cha}$ denote the spatial-wise multiplication. As a result, channel and spatial information from pre- and post-disaster branches are effectively aggregated in the cross-directional fusion model.
The proposed fusion model consists of only simple convolution and matrix operations, which is easy to implement and integrate with existing CNN architectures.

\begin{figure}
  \centering
  \includegraphics[width=0.8\linewidth]{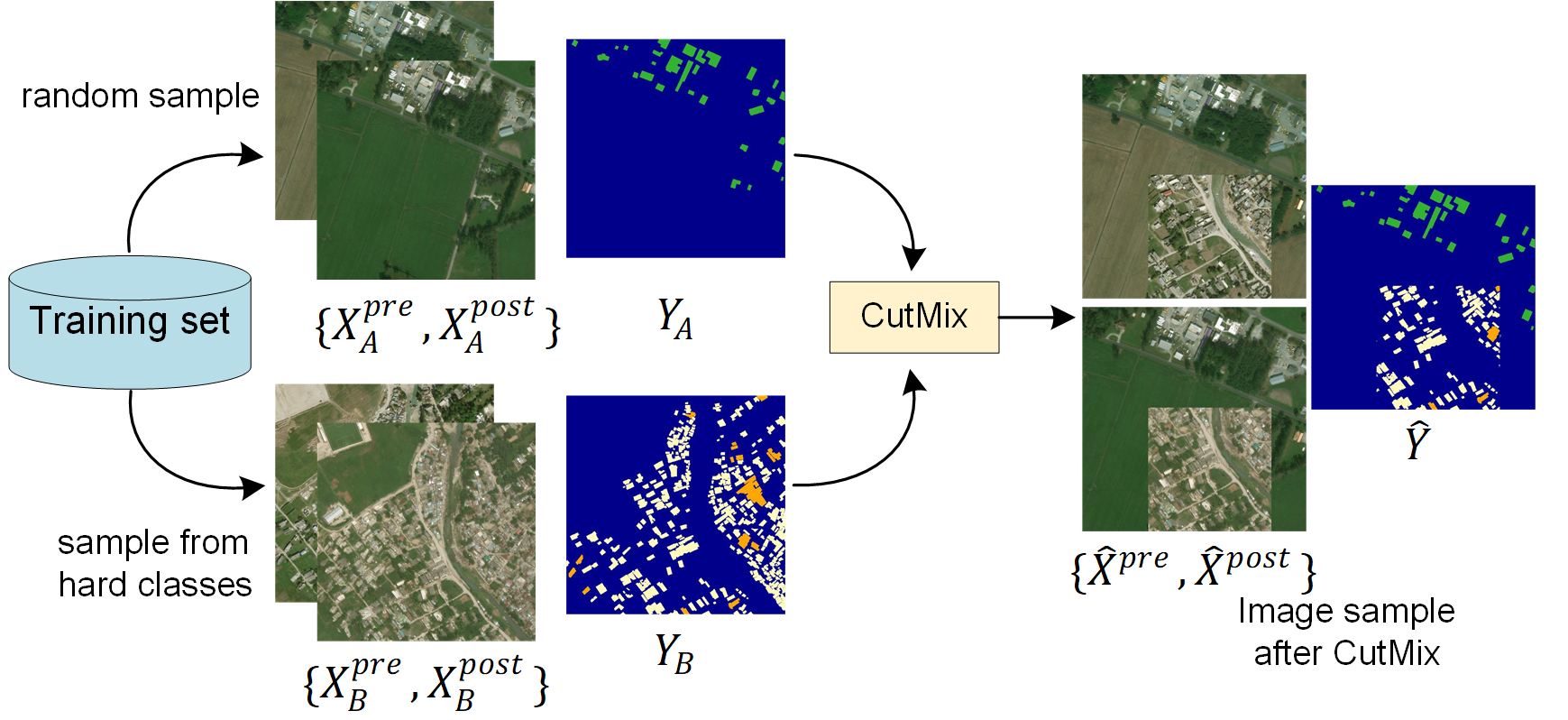} \\
   \includegraphics[width=0.48\linewidth]{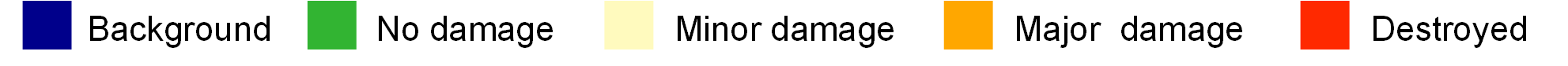}
  \caption{Data augmentation with CutMix for hard classes.}
  \vspace{-0.1cm}
  \label{figCutmix}
\end{figure}

\textbf{Data augmentation with CutMix.} 
As discussed in Sec. \ref{secIntro}, there are several damage levels that are difficult to distinguish from each other due to object visual similarities in satellite images. To address this challenge, we leverage the CutMix data augmentation scheme to increase the sample sizes of hard classes, hoping to build better feature representations for these classes.  
Specifically, image patches are cut from samples that contain hard classes (i.e., \textit{minor} and \textit{major} damage classes for the xBD dataset based on the results in Table \ref{tabConfusion}), and then pasted into any random sample images. The CutMix procedure is illustrated in Fig. \ref{figCutmix}. Let $\{X_A^{pre}, X_A^{post}\}  \in \mathbb{R}^{c\times H\times W}$ and $Y_A \in \mathbb{R}^{H\times W}$ be a randomly selected training sample (an image pair) and label, ($X_B^{pre}, X_B^{post}, Y_B$) be a randomly selected sample from hard classes, where $c$ is the channel number of images. Then the CutMix operation in this task can be defined as
\begin{equation}
\begin{aligned}
\hat{X}^{pre} = M\odot X_A^{pre} + (1-M) \odot X_B^{pre} , \\
\hat{X}^{post} = M\odot X_A^{post} + (1-M) \odot X_B^{post},\\
\quad \hat{Y} = M\odot Y_A + (1-M) \odot Y_B,  
\end{aligned}
\label{eqConv1}
\end{equation}
where $M\in \{ 0,1\}^{W\times H}$ denotes a binary mask indicating where to cut out and fill in from two image samples, $\odot$ is an element-wise multiplication, and $(\hat{X}^{pre}, \hat{X}^{post}, \hat{Y})$ represents the generated new sample. With the increased sample sizes of hard classes using CutMix, the network is forced to pay more attention to these classes and learn more robust representations for them.

\section{Experiments}
\begin{wraptable}{r}{0.3\textwidth}
\vspace{-\intextsep}
\footnotesize
  \caption{xBD dataset splits and annotation numbers.}
  \vspace{-0.3cm}
  \label{tabDataset}
  \begin{center}
  \setlength\tabcolsep{2.0pt} 
  \renewcommand{\arraystretch}{0.8}
  \begin{tabular}{ccc}
    \toprule
    {Split}     & {Image No.}     & {Polygons No.} \\
    \midrule
    Train & 18336  &  632228     \\
    Test    & 1866  & 109724     \\
    \bottomrule
  \end{tabular}
  \vspace{-0.3cm}
  \end{center}
\end{wraptable}
\textbf{Dataset description.} The xBD dataset is a large-scale public dataset of satellite images for building segmentation and damage assessment \cite{gupta2019xbd}. It covers a variety of disasters (such as hurricanes, floods, wildfire and earthquakes) and locations with more than 800,000 building annotations across the world. The dataset consists of image pairs (pre- and post-disaster) with a size of $1024 \times 1024$ pixels. The damage assessment contains 4 levels, including no damage, minor damage, major damage and destroyed. The training and testing sets are listed in Table \ref{tabDataset}. It is worth mentioning that the data is imbalanced and the damage level is highly skewed toward ``no damage''. The number of each damage level's polygons is reported in Table \ref{tabDamageLevel}.
\begin{wraptable}{r}{0.4\textwidth}
\vspace{-\intextsep}
\footnotesize
  \caption{Damage level annotations.}
  \vspace{-0.3cm}
  \begin{center}
  \setlength\tabcolsep{2.0pt} 
  \renewcommand{\arraystretch}{0.7}
  \begin{tabular}{ccccc}
    \toprule
    \textbf{}     & No damage    & Minor & Major & Destroyed \\
    \midrule
   \textbf{No.} & 313003  &  36860    & 29904 & 31560 \\
    \bottomrule
  \end{tabular}
  \vspace{-0.3cm}
  \end{center}
    \label{tabDamageLevel}
\end{wraptable}

All experiments are evaluated on xBD dataset with metric $F1_b$ for building segmentation, which is defined as:
\begin{equation}
F1_b = \frac{2TP}{2TP+FP+FN},
\label{eqF1}
\end{equation}
where $TP$, $FP$ and $FN$ are the number of true-positive, false-positive and false-negative pixels of segmentation results. The $F1_d$ metric for damage assessment is in a similar manner with $F1_b$. The overall score $F1_s$ of building segmentation and damage assessment is defined as:
\begin{equation}
F1_s = 0.3 \times F1_b + 0.7 \times F1_d.
\label{eqOveral}
\end{equation}

\textbf{Implementation details.}
We use Pytorch framework to build the networks. All experiments are conducted on a machine with an Intel i9-9920X CPU and two NVIDIA TITAN-V GPUs. All images are cropped to a size of $512\times 512$ pixels for training. Apart from CutMix, basic data augmentation is used, such as flip and rotation. The cross-entropy loss is used for both building segmentation and damage assessment. The optimization method is Adam. In the building segmentation stage, the learning rate is 0.00015 and the number of epoch is 120. In the building damage assessment stage, the learning rate is 0.0002 and the number of epoch is 20.

\textbf{Results analysis.}
To validate the effectiveness of the proposed framework, we employ state-of-the-art methods for comparison on xBD. All the models adopt Res50 as backbone. A Res50 baseline without cross-directional fusion and CutMix is also used for comparison. As shown in Table \ref{tabResults}, the proposed framework using the vanilla U-Net structure alone (i.e., Res50 (baseline)) is able to outperform the existing methods in terms of all three metrics. With the proposed cross-directional fusion model and CutMix data augmentation, it further improves the accuracy for damage assessment (0.778 vs. 0.757 in $F1_d$), and the improvement is consistent for all the damage levels. 
Several visual examples of building segmentation and damage assessment results of our method are presented in the \textbf{Appendix}. 
\begin{table}
  \caption{Building segmentation and damage assessment performance comparison on xBD dataset.}
  \footnotesize
  \centering
  \begin{tabular}{cccccccc}
    \toprule
         & $F1_s$ (overall) & $F1_b$ & $F1_d$  & No damage    & Minor & Major & Destroyed \\
    \midrule
    Gupta et al. \cite{gupta2020rescuenet} & 0.741 & 0.835 &	0.697 &	0.906 &	0.493 &	0.722 &	0.837 \\
   Weber et al. \cite{weber2020building}  & 0.770 &	0.840 &	0.740 &	0.885 &	0.563 &	0.771 &	0.808   \\
   Res50 (baseline) & 0.789 & 0.864 &	0.757 &	0.923 &	0.578 &	0.760 &	0.869  \\
   Res50 (ours) &  \textbf{0.804} &	0.864 &	0.778 &	0.927 &	0.610 &	0.781 &	0.873   \\
    \bottomrule
  \end{tabular}
    \label{tabResults}
\end{table}

\begin{wraptable}{r}{0.37\textwidth}
\vspace{-\intextsep}
\footnotesize
  \caption{Parameters (M) and FLOPs (G) of our method and baseline.}
  \vspace{-0.3cm}
  \begin{center}
  \setlength\tabcolsep{2.0pt} 
  \renewcommand{\arraystretch}{0.4}
  \begin{tabular}{cccc}
    \toprule
    \textbf{Method}     & Params    & FLOPs  \\
    \midrule
     Res50 (baseline)   & 32.5  & 92.9  \\
     Ours    & 33.1  &  107.6     \\
    \bottomrule
  \end{tabular}
  \vspace{-0.3cm}
  \end{center}
\label{tabParam}
\end{wraptable}

We also compare the parameter size and computational cost (FLOP) between the proposed method and the baseline. As shown in Table \ref{tabParam}, although the cross-directional fusion model is added in the network, it brings less than 1M additional parameters. And this results in only slight increase in computational cost (107.6 vs. 92.9 GFLOPs). 
\begin{wraptable}{r}{0.7\textwidth}
\vspace{-\intextsep}
\footnotesize
  \caption{Ablation study on cross-directional fusion and CutMix.}
  \vspace{-0.3cm}
  \begin{center}
  \setlength\tabcolsep{2.0pt} 
  \renewcommand{\arraystretch}{0.7}
  \begin{tabular}{cccccccc}
    \toprule
    \textbf{Method}     & $F1_s$    & $F1_b$ & $F1_d$ & No damage & Minor & Major & Destroyed \\
    \midrule
     Res50 (baseline) & 0.789 & 0.864 &	0.757 &	0.923 &	0.578 &	0.760 &	0.869 \\
     Ours w/o CutMix   & 0.795  &  0.864    & 0.765  & 0.923 & 0.592 & 0.769 &0.871\\
     Ours w/o Fusion   & 0.802  &  0.864    & 0.775  & 0.926 & 0.605 & 0.779 & 0.872 \\
    \bottomrule
  \end{tabular}
  \vspace{-0.3cm}
  \end{center}
\label{tabAbfusion}
\end{wraptable}

\textbf{Ablation study.}
To investigate the contribution of cross-directional fusion model and CutMix data augmentation, we perform ablation studies on these two components. From Table \ref{tabAbfusion}, we can observe that the cross-directional fusion model brings 0.6\% (0.789 vs. 0.795 in $F1_s$) improvement compared with the baseline. Moreover, the CutMix strategy on hard classes boosts the overall performance ($F1_s$) by 1.3\%, which is a considerable gain. In particular, the accuracy of \textit{minor damage} class is improved by 2.7\% (0.578 vs. 0.605) with CutMix. 

\begin{wraptable}{r}{0.5\textwidth}
\vspace{-\intextsep}
  \caption{Ablation study of CutMix on different classes.}
  \vspace{-0.2cm}
  \footnotesize
  \centering
  \setlength\tabcolsep{2.0pt} 
  \renewcommand{\arraystretch}{0.7}
  \begin{tabular}{cccccc}
    \toprule 
         Backbone & No damage & Minor & Major & Destroyed  & $F1_s$ \\
    \midrule
    Res50 & & & & & 0.789  \\
    Res50 & \checkmark  & \checkmark &  \checkmark  &   \checkmark  & 0.790\\
    Res50 &  & \checkmark &  \checkmark  &   \checkmark  & 0.797\\
    Res50 &   & \checkmark &  \checkmark  &     & \textbf{0.802}\\
    Res50 &   & \checkmark &    &     & 0.795\\
    \bottomrule
  \end{tabular}
 \label{tabAbcut}
\end{wraptable}
For the xBD dataset, \textit{minor damage} and \textit{major damage} are the most difficult classes based on the results in Table \ref{tabAbfusion}. Therefore, in our proposed method, CutMix data augmentation is mainly focused on these two damage levels to generate more training samples for them. We further conduct experiments to investigate the influence of CutMix on different damage levels. Table \ref{tabAbcut} reports the overall scores when CutMix is applied on different damage levels. We can clearly observe that there is barely any improvement when CutMix is used on the entire dataset (i.e., considering all classes). In contrast, a significant improvement is achieved when only \textit{minor} and \textit{major} damage levels are considered in the data augmentation. This hard-class biased data augmentation strategy facilitates the network to learn better representations for those classes. 
\vspace{-0.2cm}

\section{Conclusion}
In this work, we propose a two-stage U-Net framework that integrates cross-directional fusion model for building damage assessment. In the fusion model, channel and spatial information from pre- and post-disaster image features is aggregated and embedded in the network, which enables the network to learn feature representations more effectively. Moreover, a data augmentation strategy CutMix is used to mitigate the challenge of hard classes. Experimental results show that significant improvements can be achieved when CutMix is applied on hard damage levels. The proposed method also yields state-of-the-art building damage assessment performance on the xBD dataset.


\bibliographystyle{unsrt}
\bibliography{neurips}






\clearpage
\appendix
\section{Appendix}

\subsection{Visual Examples}
\begin{figure}[h]
	\centering
    \subfloat{
		\includegraphics[width=0.22\linewidth]{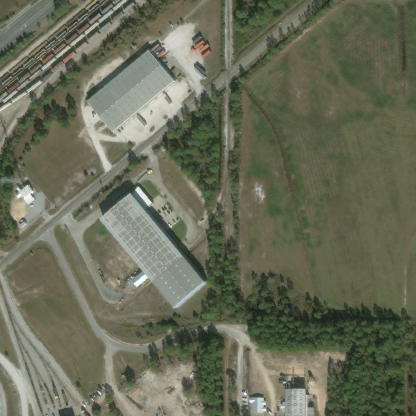}
	} 
	\subfloat{
		\includegraphics[width=0.22\linewidth]{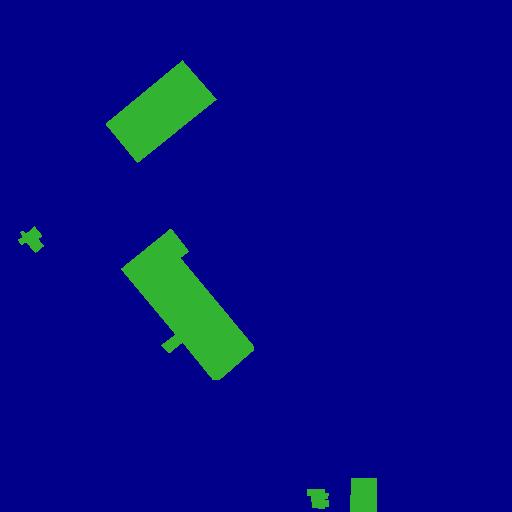}
    }  
    \subfloat{
		\includegraphics[width=0.22\linewidth]{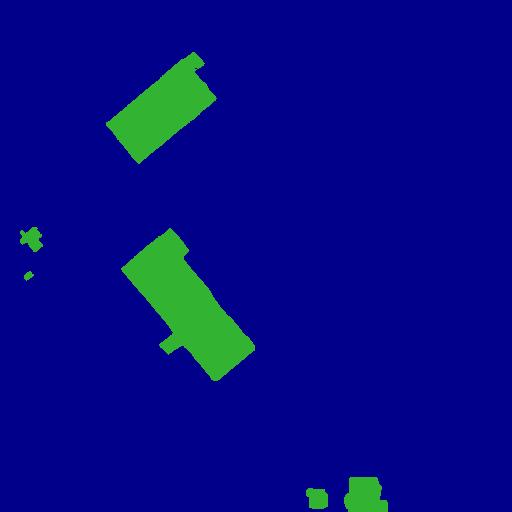}
    }  
    \subfloat{
		\includegraphics[width=0.22\linewidth]{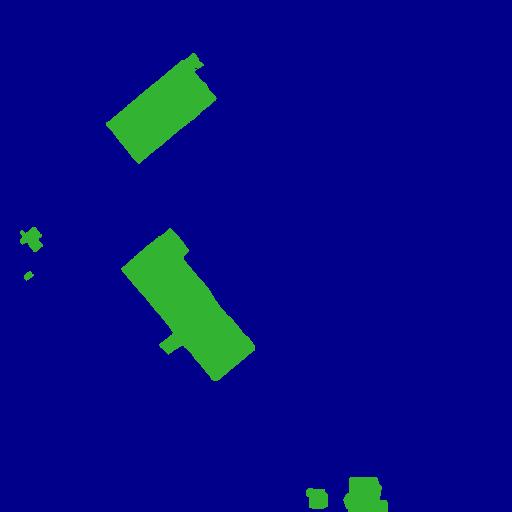}
    } \\
    \vspace{-0.2cm}
    \subfloat[pre and post-disaster]{
		\includegraphics[width=0.22\linewidth]{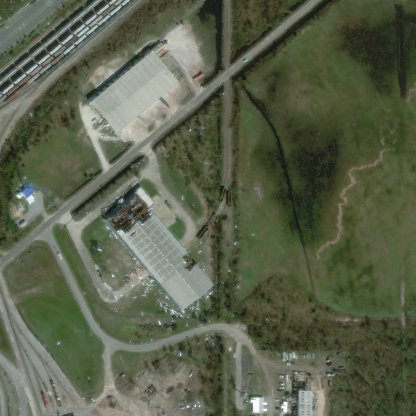}
	} 
	\subfloat[ground-truth]{
		\includegraphics[width=0.22\linewidth]{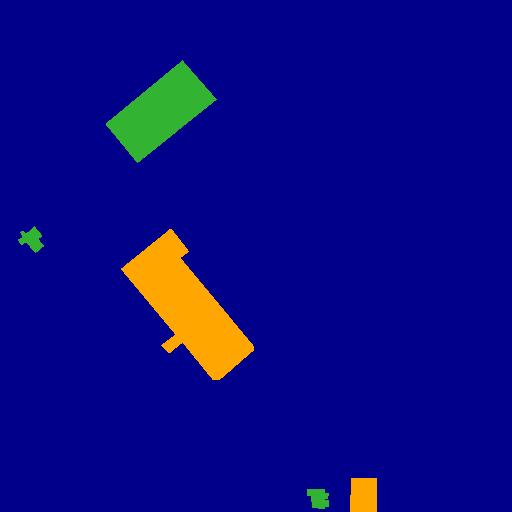}
    }  
    \subfloat[baseline]{
		\includegraphics[width=0.22\linewidth]{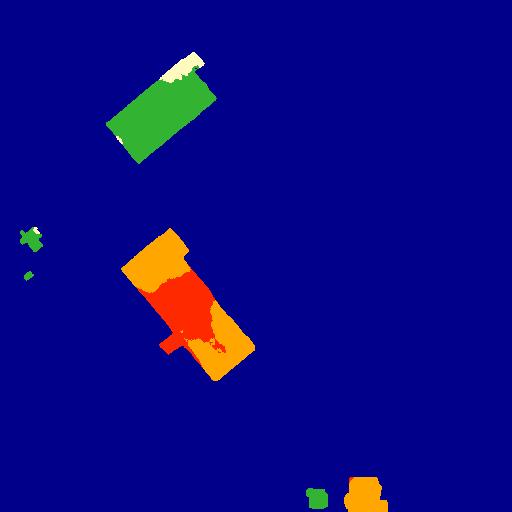}
    }  
    \subfloat[ours]{
		\includegraphics[width=0.22\linewidth]{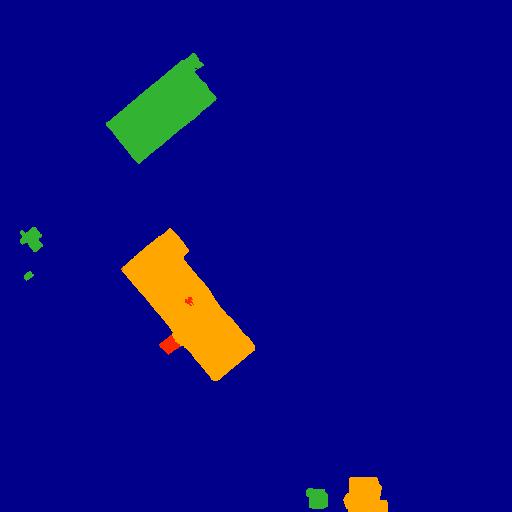}
    }  \\
    \vspace{-0.2cm}
    \subfloat{
		\includegraphics[width=0.65\linewidth]{legend.png}
    } 
	\caption{A visual comparison of results of the baseline and the proposed method over a hurricane disaster image. }
\label{figExample1}
\end{figure}

\begin{figure}[h]
	\centering
	\subfloat{
		\includegraphics[width=0.22\linewidth]{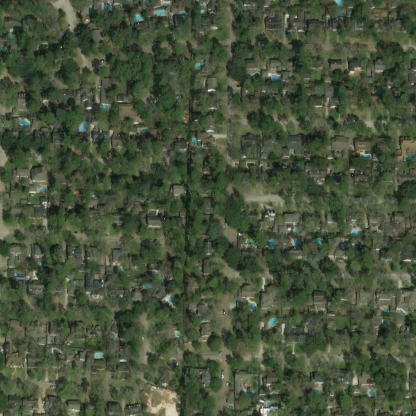}
	} 
	\subfloat{
		\includegraphics[width=0.22\linewidth]{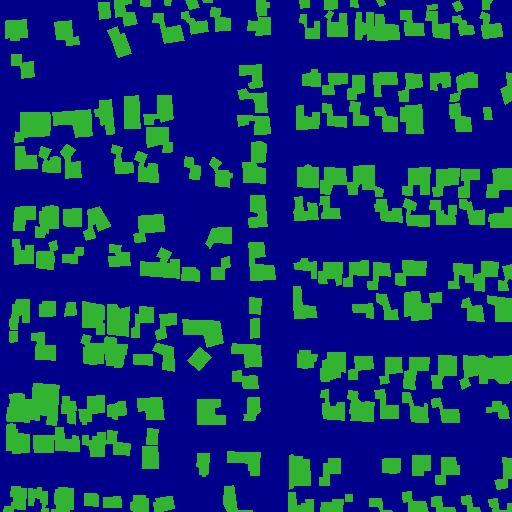}
    }  
    \subfloat{
		\includegraphics[width=0.22\linewidth]{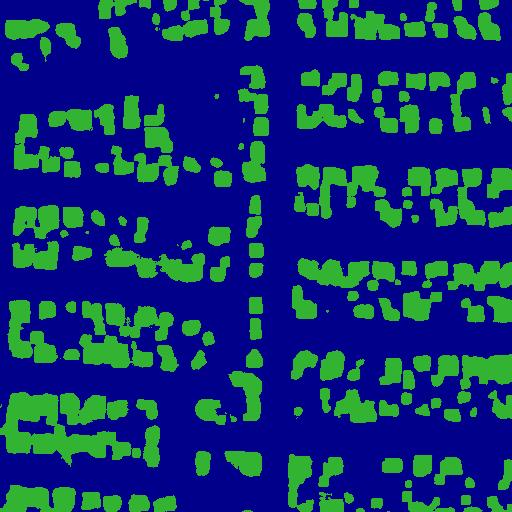}
    }  
    \subfloat{
		\includegraphics[width=0.22\linewidth]{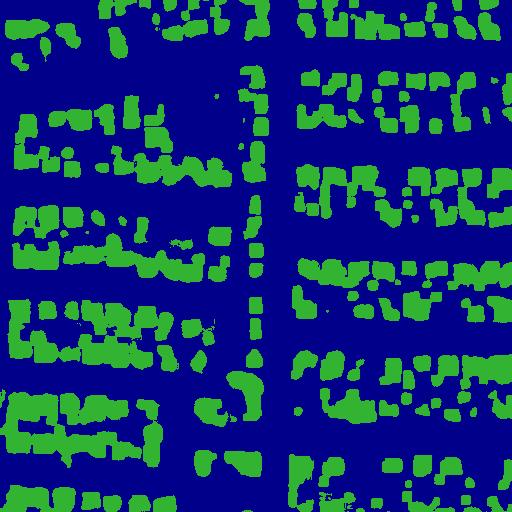}
    } \\ 
    \vspace{-0.2cm}
    \subfloat[pre and post-disaster]{
		\includegraphics[width=0.22\linewidth]{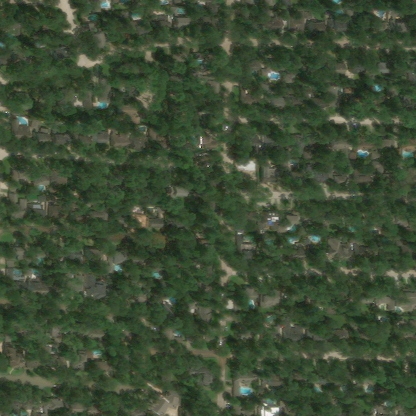}
	} 
	\subfloat[ground-truth]{
		\includegraphics[width=0.22\linewidth]{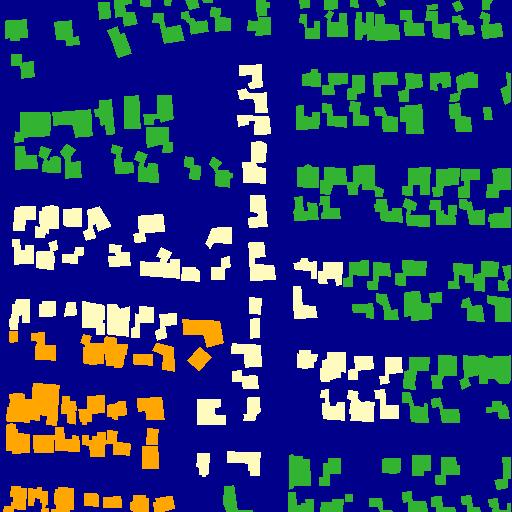}
    }  
    \subfloat[baseline]{
		\includegraphics[width=0.22\linewidth]{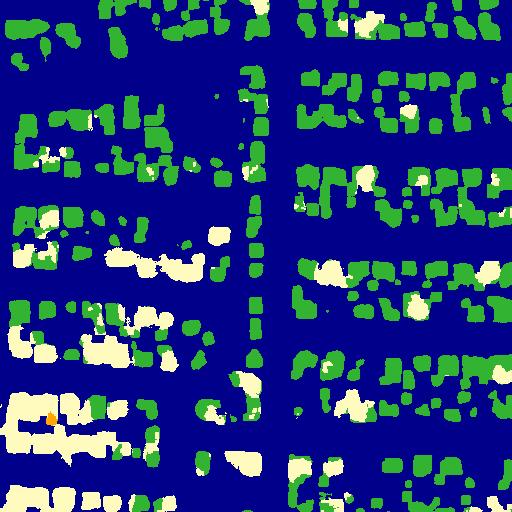}
    }  
    \subfloat[ours]{
		\includegraphics[width=0.22\linewidth]{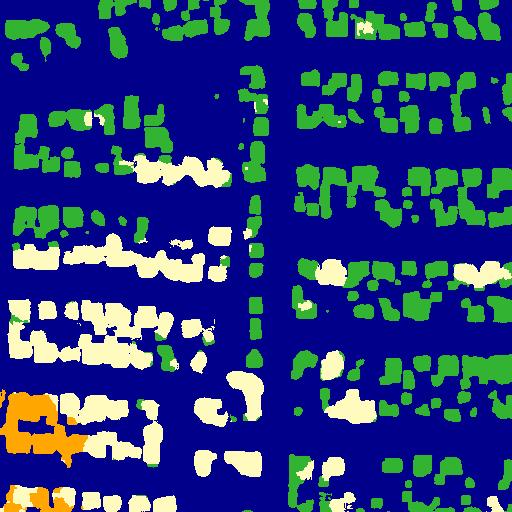}
    }  \\
    \vspace{-0.2cm}
    \subfloat{
		\includegraphics[width=0.65\linewidth]{legend.png}
    } 
	\caption{A visual comparison of results of the baseline and the proposed method over a hurricane disaster image.}
\label{figExample2}
\end{figure}

\begin{figure}
	\centering
    \subfloat{
		\includegraphics[width=0.22\linewidth]{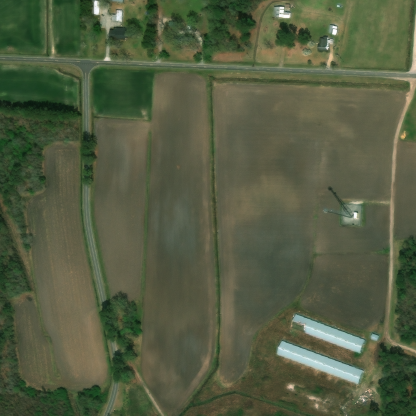}
	} 
	\subfloat{
		\includegraphics[width=0.22\linewidth]{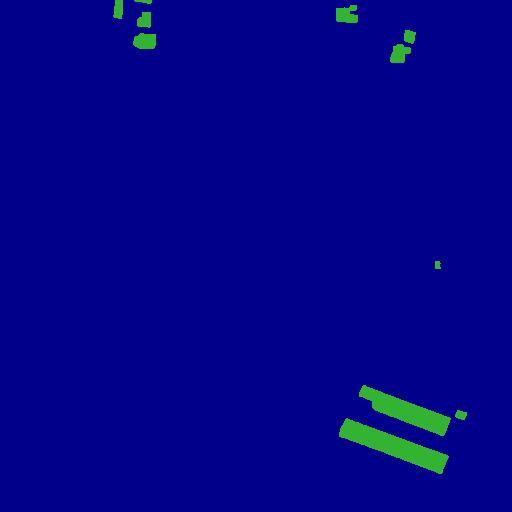}
    }  
    \subfloat{
		\includegraphics[width=0.22\linewidth]{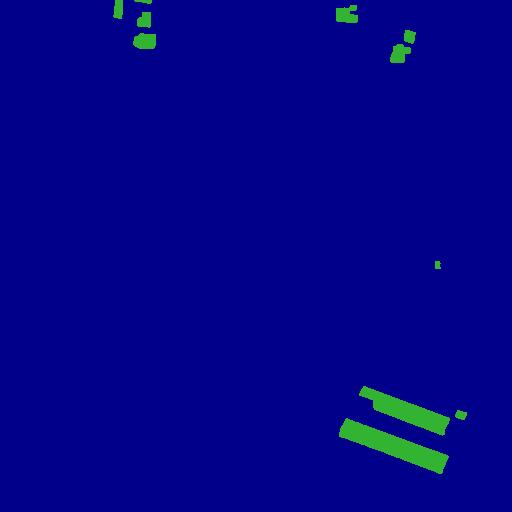}
    }  
    \subfloat{
		\includegraphics[width=0.22\linewidth]{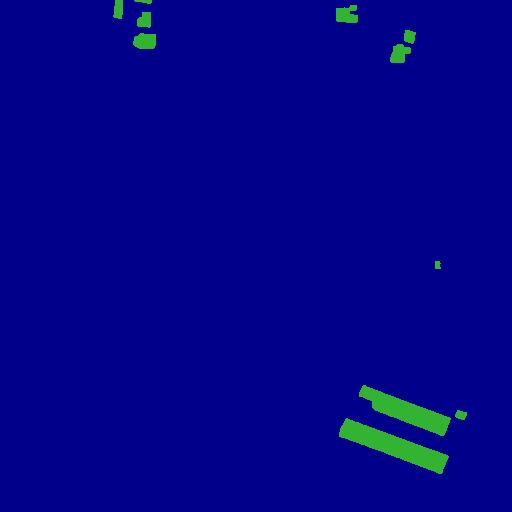}
    } \\ 
    \vspace{-0.2cm}
    \subfloat[pre and post-disaster]{
		\includegraphics[width=0.22\linewidth]{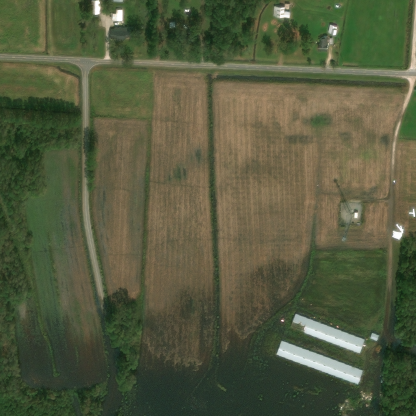}
	} 
	\subfloat[ground-truth]{
		\includegraphics[width=0.22\linewidth]{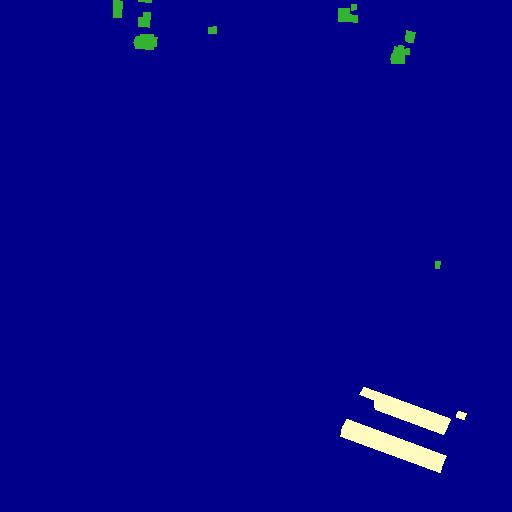}
    }  
    \subfloat[baseline]{
		\includegraphics[width=0.22\linewidth]{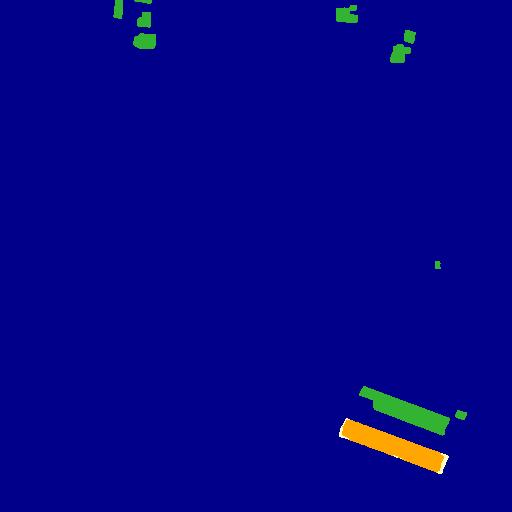}
    }  
    \subfloat[ours]{
		\includegraphics[width=0.22\linewidth]{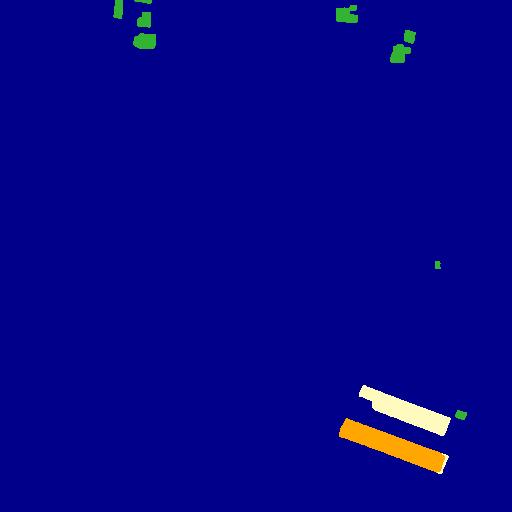}
    }  \\
    \vspace{-0.2cm}
    \subfloat{
		\includegraphics[width=0.65\linewidth]{legend.png}
    } 
	\caption{A visual comparison of results of the baseline and the proposed method over a tsunami disaster image. }
\label{figExample3}
\end{figure}

\begin{figure}
	\centering
    \subfloat{
		\includegraphics[width=0.22\linewidth]{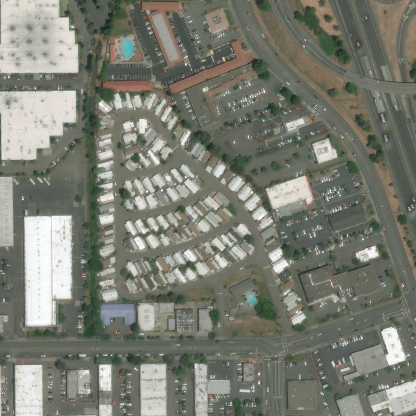}
	} 
	\subfloat{
		\includegraphics[width=0.22\linewidth]{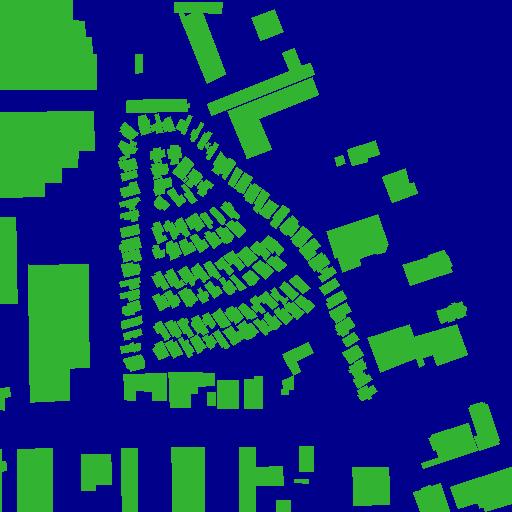}
    }  
    \subfloat{
		\includegraphics[width=0.22\linewidth]{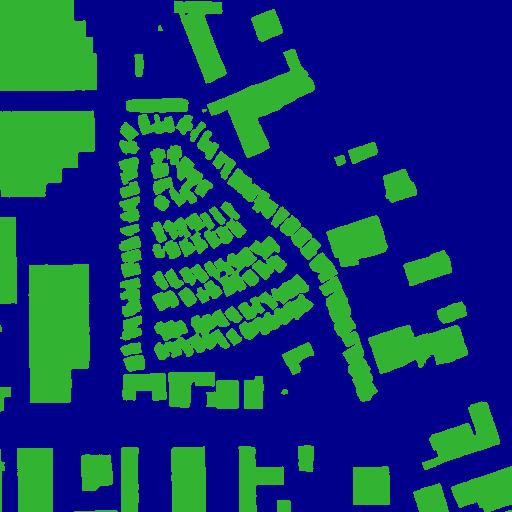}
    }  
    \subfloat{
		\includegraphics[width=0.22\linewidth]{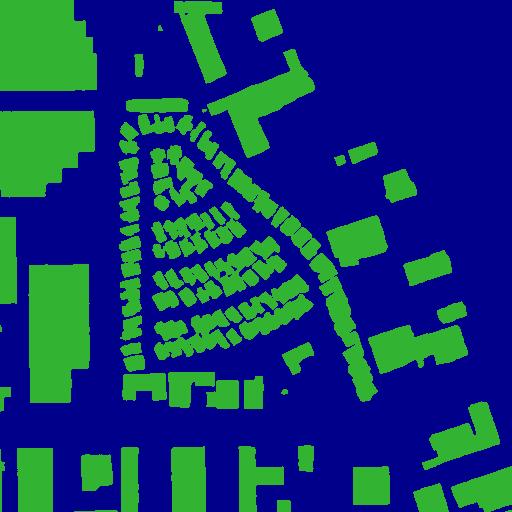}
    } \\ 
    \vspace{-0.2cm}
    \subfloat[pre and post-disaster]{
		\includegraphics[width=0.22\linewidth]{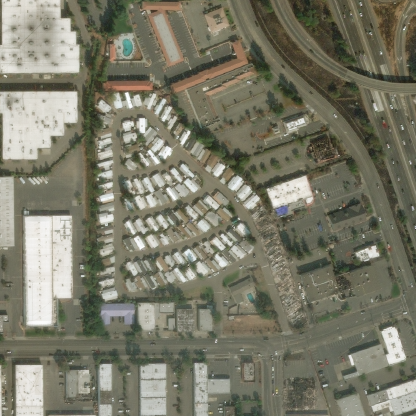}
	} 
	\subfloat[ground-truth]{
		\includegraphics[width=0.22\linewidth]{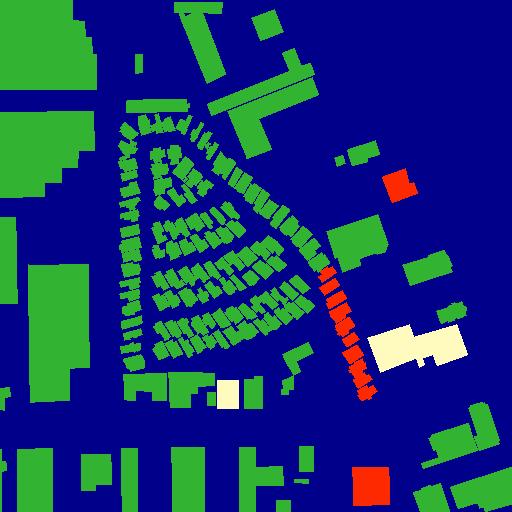}
    }  
    \subfloat[baseline]{
		\includegraphics[width=0.22\linewidth]{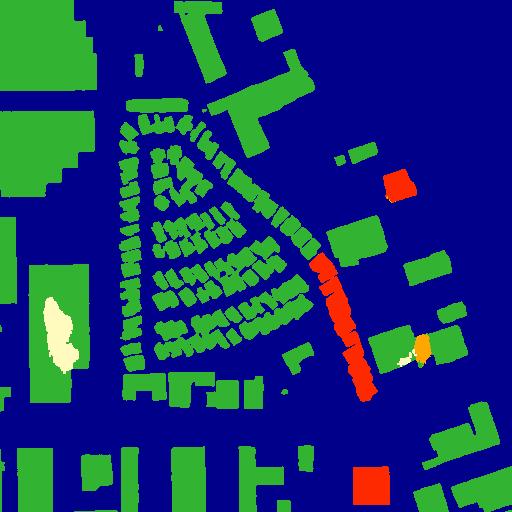}
    }  
    \subfloat[ours]{
		\includegraphics[width=0.22\linewidth]{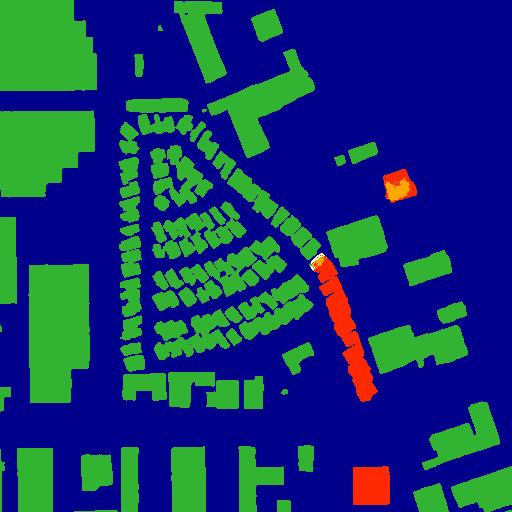}
    }  \\
    \vspace{-0.2cm}
    \subfloat{
		\includegraphics[width=0.65\linewidth]{legend.png}
    } 
	\caption{A visual comparison of results of the baseline and the proposed method over a wildfire disaster image. }
\label{figExample4}
\end{figure}

Figures \ref{figExample1}-\ref{figExample4} provide visual comparisons of the building damage assessment results of our proposed method and the baseline for different disasters. We can observe that our proposed framework achieves better performance in damage level assessment. For example, in Figure \ref{figExample1}, some regions of \textit{major damage} (orange color) and \textit{no damage} (green color) are mis-classified as \textit{destroyed} (red color) and \textit{minor damage} (yellow color) respectively by the baseline, while our proposed method yields much less mis-classification and produces a segmentation map that is very close to the ground-truth.

\end{document}